\documentclass[sigconf]{acmart}
\AtBeginDocument{%
  }

\setcopyright{none}
\acmConference[KDD Workshop '25]{Workshop on Structured Knowledge for Large Language Models (SKnowLLM) at KDD 2025}{August 03--07,2025}{Toronto, Canada}

\usepackage{times}
\usepackage{latexsym}

\usepackage[T1]{fontenc}

\usepackage[utf8]{inputenc}

\usepackage{microtype}

\usepackage{graphicx}

\usepackage{amsmath}
\usepackage{booktabs}
\usepackage{subcaption}
\usepackage{enumitem}

\usepackage{listings}
\begin{document}

\title{Multi-task retriever fine-tuning for domain-specific and efficient RAG}

\author{Patrice Bechard}
\affiliation{%
    \institution{ServiceNow}
    \city{Montreal}
    \country{Canada}
}
\email{patrice.bechard@servicenow.com}

\author{Orlando Marquez Ayala}
\affiliation{%
    \institution{ServiceNow}
    \city{Montreal}
    \country{Canada}    
}
\email{orlando.marquez@servicenow.com}


\begin{abstract}
Retrieval-Augmented Generation (RAG) has become ubiquitous when deploying Large Language Models (LLMs), as it can address typical limitations such as generating hallucinated or outdated information. However, when building real-world RAG applications, practical issues arise. First, the retrieved information is generally domain-specific. Since it is computationally expensive to fine-tune LLMs, it is more feasible to fine-tune the retriever model to improve the quality of the data included in the LLM input. Second, as more applications are deployed in the same real-world system, one cannot afford to deploy separate retrievers for different tasks. Moreover, these RAG applications normally retrieve different kinds of data. Our solution is to instruction fine-tune a small retriever model on a variety of domain-specific tasks to allow us to deploy one model that can serve many use cases, thereby achieving low-cost, scalability, and speed. We show how this model generalizes to out-of-domain settings as well as to an unseen retrieval task on real-world enterprise use cases.
\end{abstract}



\received{30 May 2025}

\settopmatter{printacmref=false} 

\maketitle

\section{Introduction}
\label{sec:intro}

As more and more Generative AI (GenAI) applications are integrated into real-world production systems, Retrieval-Augmented Generation (RAG) has been adopted in industry as a common technique to improve the output of Large Language Models (LLMs). RAG alleviates inherent LLM pitfalls such as propensity to hallucinate, generating outdated knowledge, and lack of traceability to data sources \cite{10.1145/3637528.3671470, gao2024retrievalaugmentedgenerationlargelanguage}. 

Introducing a retrieval step into the generation process introduces, however, several practical challenges. While an LLM with a large number of parameters, such as GPT-4 \cite{openai2024gpt4technicalreport}, can be prompted to work with any kind of input and generate any kind of textual output, the retriever needs to be small, fast, and perform well with data sources that tend to be domain-specific.

\begin{figure}[t]
    \centering
    \includegraphics[width=\linewidth]{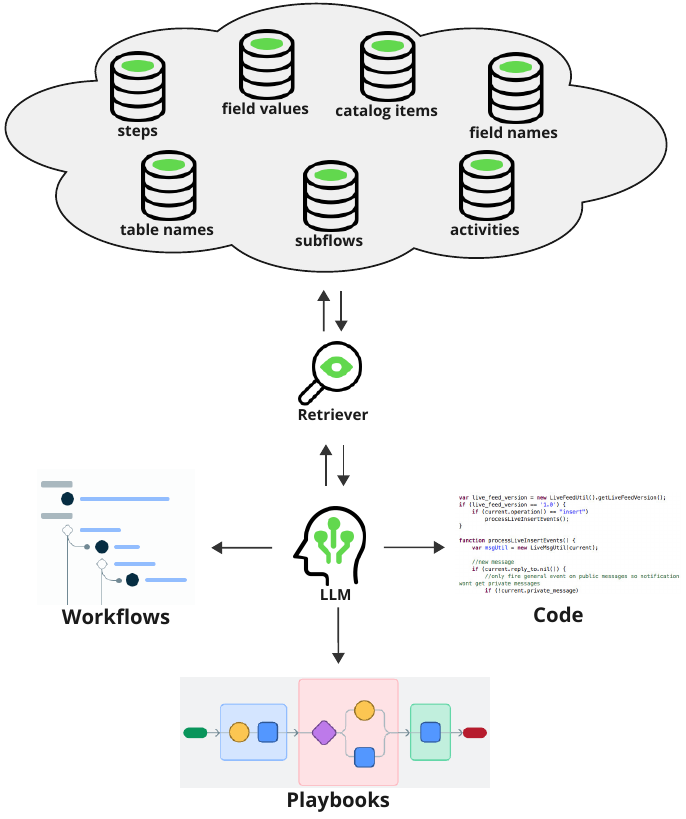}
    \caption{Given an ecosystem of RAG applications, how do we build a retriever that can adapt to a specific domain and to a variety of retrieval tasks?}
    \Description{Architecture of structured data retrieval for multiple tasks.}
    \label{fig:problem_defin}
\vspace*{-0.5cm}
\end{figure}

Off-the-shelf retrievers of different sizes are available to AI practitioners. Embedding services such as Voyage\footnote{https://docs.voyageai.com/docs/embeddings} perform well on open-source benchmarks but they do not necessarily generalize to the kind of data seen in real-world settings, especially when this data is structured and comes from existing databases.

Another practical challenge is achieving scalability and generalization across different GenAI use cases that depend on retrieval. A crucial advantage of LLMs compared to traditional machine learning models is that they can generalize to a myriad of tasks due to vast amounts of pretraining data and instruction fine-tuning \cite{wei2022finetuned, zhang2024instructiontuninglargelanguage, NEURIPS2022_b1efde53}. But if the retriever does not perform well and fast across many retrieval tasks, the downstream generation will be negatively affected.

The problem we are trying to solve is then: \textit{how to adapt the retrieval step to a specific domain and to a variety of retrieval tasks?} In this work, we are not interested in the choice of LLM, assuming that improvements in the retrieved results translate into improvements in the downstream generation task.

The context is an enterprise company that deploys several GenAI applications that currently rely or will rely on RAG: Flow Generation \cite{bechard2024reducing}, Playbook Generation\footnote{www.servicenow.com/docs/bundle/xanadu-intelligent-experiences/page/administer/now-assist-platform/concept/now-assist-playbook-generation-skill.html}, and Code Generation. Workflows are step-by-step processes that automate one or more goals while playbooks contain workflows and other UI components such as forms. The objective of these applications is to generate domain-specific workflows, playbooks, and code from textual input.

Our solution is to instruction fine-tune a small retriever on a variety of tasks. This retriever is deployed in the ecosystem of GenAI applications, which prompt it to retrieve desired structured data from databases. Information such as workflow step names, table names, and field names are then passed to LLMs to generate workflows or playbooks, leading to higher output quality.

To build a multi-task retriever dataset, we extracted data from internal databases and reused the Flow Generation training set. For our solution, we fine-tune mGTE \cite{zhang-etal-2024-mgte} because of its large context length (8,192 tokens), allowing it to receive long instructions, and because of its multilingual capabilities. We compare it with BM25, a simple yet powerful term-frequency method \cite{10.5555/188490.188561}, and with recent open-source multilingual embedding models: mE5 \cite{wang2024multilinguale5textembeddings} and mGTE. We also include evaluation of larger instruction-tuned models based on LLMs, namely GTE-Qwen-2 \cite{yang2024qwen2}, E5-mistral \cite{wang-etal-2024-improving-text}, and GritLM \cite{muennighoff2024generative}.

We perform several evaluations. First, we evaluate on the tasks that the retriever was trained on, but on out-of-domain (OOD) settings. The internal training datasets come from the IT domain, but the OOD splits come from diverse domains such as HR and finance. We then evaluate on a related but different retrieval task to test the generalization ability of the model. For example, while the model was trained to retrieve step names based on text, we also want to retrieve relevant workflow structures based on text. This is a related task because workflows are made up of steps. Lastly, to see whether the multilingual abilities of models are preserved after our multi-task fine-tuning, we evaluate on input from different languages even though our retrieval training dataset is only in English.


Our contributions are the following:
\begin{itemize}
\itemsep-0.2em
    \item We provide a case study for how to build a domain-specific and efficient retriever for real-world RAG.
    \item We demonstrate that multi-task retriever fine-tuning can lead to generalization to out-of-domain datasets and to related but different retrieval tasks.
\end{itemize}

\section{Related Work}
\label{sec:related_work}

There have been several efforts to achieve \textbf{domain-specific RAG}. While more complex, jointly training the retriever and the generator has been shown to work well for question-answering \cite{siriwardhana-etal-2023-improving}. Others encode the domain-specific information in knowledge graphs in addition to vector databases \cite{barron2024domainspecificretrievalaugmentedgenerationusing}. Retrieval-Augmented Fine Tuning \cite{zhang2024raft} helps the LLM adapt to the domain by modifying how the retrieved information is present in the LLM training dataset. In contrast, our approach relies solely on training a better retriever, which can be used with any LLM, thereby reducing coupling.

\textbf{Multi-task retriever models} \cite{maillard2021multi, wang2022text} enhance embedding versatility by simultaneously training on multiple tasks or by utilizing a large dataset spanning diverse topics. Building upon this concept, \textbf{instruction-following embedding models} \cite{asai2023task, su2022one, behnamghader2024llm2vec} integrate explicit instructions into the embedding process, thereby enabling the generation of more nuanced and task-specific embeddings. We leverage this kind of instruction fine-tuning to train a retriever on many datasets containing structured data extracted from databases.

Lastly, generating workflows is a structured output task, similar to code generation, that requires specialized embeddings. \textbf{Code embedding models} \cite{feng2020codebert, li2022coderetriever, guo2020graphcodebert, neelakantan2022text, zhang2024codesage} adapt common training techniques to the domain of code representation learning and retrieval. A parallel research trajectory focuses on \textbf{structured data embedding models}, including Synchromesh \cite{poesia2022synchromesh}, which fine-tunes a model to retrieve relevant samples for few-shot prompting using a similarity metric derived from tree edit distance, and SANTA \cite{li2023ge}, which implements a modified pretraining objective to generate structure-aware representations. Prior work already uses RAG for a similar but more limited structured output task  \cite{bechard2024reducing}. This paper is a natural extension to increase reusability of the retriever in an expanding ecosystem of GenAI applications.

\section{Methodology}
\label{sec:methodology}

To build a multi-task retriever, we first had to define the tasks and then build their datasets, keeping in mind that we could not do any labeling for these tasks. The starting point was to define the data that the existing RAG applications would need.

The GenAI applications currently deployed in our system require diverse types of retrieved data to generate output acceptable to users. As these applications become more sophisticated, the extent and type of data retrieved can only increase. But currently we focus mostly on \textit{steps}, which are used by workflows and playbooks, and \textit{tables}, which are used by workflows and playbooks, and can be used in code generation. Steps are building blocks of processes; they can be \textit{actions}, \textit{subflows}, or \textit{activities}. When it comes to tables, the crucial information are \textit{table names} and \textit{table field names}. For instance, a user may want to generate code that makes a database call, but without grounding the LLM on actual table information, the LLM may refer to a non-existent table name.

\begin{figure}[htbp]
  \centering
  \includegraphics[width=\linewidth]{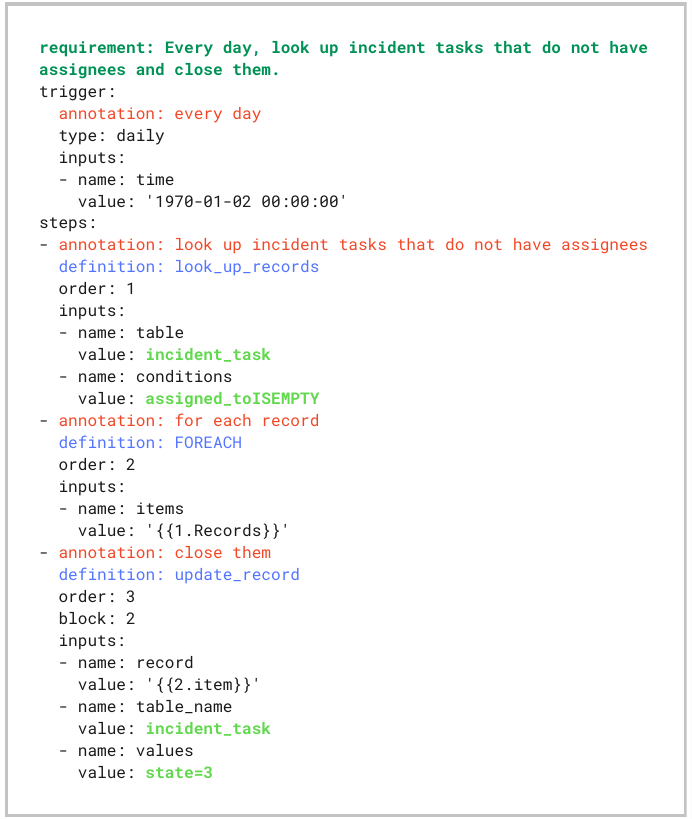}
\caption{Dataset example of \textit{Flow Generation}. We constructed retrieval multi-task examples from them.}
\Description{Example of workflow JSON}
\label{fig:sample_flow_generation}
\end{figure}

We extracted data from two sources to create the multi-task retrieval dataset. Complex and semantically diverse examples were created from the Flow Generation training set. Figure \ref{fig:sample_flow_generation} shows an example from this set in YAML format, which includes step names (\texttt{definition}), table names, and table field names (part of \texttt{conditions} and \texttt{values}). The other source are database tables that include other elements we are interested in, besides steps and tables, that contain a text field such as \textit{description}.

\subsection{Tasks}\label{sec:tasks}
From the Flow generation training set, we constructed three categories of tasks, constituting around half of all retrieval examples:
\begin{enumerate}
    \item Retrieve \textbf{steps} that can be used in workflows or playbooks.
    \item Retrieve \textbf{tables} that can be used in step inputs or in code.
    \item Retrieve \textbf{table fields} that can be used in step inputs or in code, given a table name.
\end{enumerate}

To add diversity to each of the tasks, we included several permutations of the input, such as:
\begin{itemize}
    \item Retrieve all the steps used in the workflow given the requirement. In the example in Figure \ref{fig:sample_flow_generation}, this means to retrieve \texttt{look\_up\_records}, \texttt{FOREACH}, and \texttt{update\_record} given the requirement \textit{Every day, look up incident tasks that do not have assignees and close them.}
    \item Retrieve a specific step from an annotation. Given the example in Figure \ref{fig:sample_flow_generation}, we can ask a retriever to get the step \texttt{look\_up\_records} given the text \textit{Look up incident tasks that do not have assignees}.
    \item Retrieve a table name given a workflow context. We would want the table name \textit{incident\_task} given all the previous steps (all YAML lines up to \texttt{definition: update\_record} in Figure \ref{fig:sample_flow_generation}).
    \item Retrieve a table field name from text. Given the annotation \textit{Look up incident tasks that do not have assignees}, retrieve \texttt{assigned\_to}.
\end{itemize}

The remaining half of retrieval examples come from database tables, such as \textit{catalog items}. This table represents items such as laptops that can be requested in the system. There is a field \textit{description} along the catalog item name, yielding a task where the item name is retrieved given a description.

\subsection{Dataset Generation}\label{sec:dataset_generation}

Once we defined the tasks to train the retriever, we proceeded to create a set of pairs consisting of text and objects, which can be steps, tables, table field names, catalog items, etc. The text portion of the pair is an instruction describing what has to be retrieved. Below we instruct the retriever to find steps given only a requirement.

\begin{lstlisting}[frame=single]
Represent this requirement for searching relevant steps: 
requirement: Every day, look up incident tasks that do not have assignees and close them.
\end{lstlisting}

Given that there may be more metadata available to find steps, we can add extra information to the instruction. Below we tell the retriever that this flow is part of the \texttt{teams} scope.

\begin{lstlisting}[frame=single]
Represent this flow for searching relevant steps:
type: flow 
scope: sn_ms_teams_ah 
requirement: Every day at 9am, notify me of pending tasks in a teams msg.
\end{lstlisting}

Because we instruction fine-tune the retriever, we can even add more details. For instance, the instruction can contain the workflow context. This is useful when we want the retriever to be influenced by what is already included in the workflow in addition to the last annotation.

\begin{lstlisting}[frame=single]
Represent this flow for searching relevant steps for the last step:
type: flow 
rcope: sn_ms_teams_ah 
requirement: Every day, look up incident tasks that do not have assignees and close them.
trigger:
  annotation: every day
  type: daily
  inputs:
  - name: time
  - value: '1970-01-02 00:00:00'
steps:
- annotation: look up incident tasks that do not have assignees
\end{lstlisting}

We can use similar instructions to retrieve any element. Given annotated workflows and extracts from database tables, we can therefore create a large number of samples for text/step, text/table, text/field, text/catalog item, etc. In total, we use 15 instruction templates to add variety to the input, and we plan to add more.

Following established practices in sentence embedding model training \cite{reimers2019sentence}, we create both positive and negative pairs, enabling the model to effectively discriminate and retrieve relevant elements. 

Positive samples are extracted from the labeled workflows as described above. To mine negative samples, we use two sampling strategies: 

\begin{itemize}
\item \textbf{Random negative} sampling simply takes an unrelated step, table, or field and matches it to a piece of text used in another positive sample. 
\item \textbf{Hard negative} sampling uses the \textit{structure} of the data to find harder negative examples. For example, as steps tend to be grouped in \textit{scopes}, we sample another step coming from the same scope as the positive step. Both \texttt{post\_response\allowbreak\_to\_slack} and \texttt{post\_a\_message} are in the \texttt{slack} scope, but they are used in different scenarios.
\end{itemize}

\section{Experiments}
\label{sec:experiments}
We describe the datasets used in training and evaluation, the baselines we compare against, as well as the retrieval metrics we used.

\subsection{Datasets}\label{sec:datasets}

For training, we use all pairs we could extract from the Flow generation training set and dataset tables, but for our development set, we evaluate only on step, table name, and table field retrieval, as these are the most important retrieval tasks. The development examples come only from the Flow Generation development set. Table \ref{tab:train_dev_dataset_stats} shows the number of examples for steps, tables, and fields in the multi-task development set.

\begin{table}[h]
\vspace{-0.1cm}
  \centering
  \caption{Number of training pairs and examples in the development dataset.}
\vspace{-0.2cm}
  \label{tab:train_dev_dataset_stats}
  \begin{tabular}{cccc}
    \toprule
    \textbf{Train Pairs} & \textbf{Dev Steps} & \textbf{Dev Tables} & \textbf{Dev Fields} \\
    \midrule
    172,658 & 279 & 307 & 355 \\
    \bottomrule
  \end{tabular}
\vspace{-0.2cm}

\end{table}

Although there are close to 5K unique steps in the datasets, there are around 20 that are used most frequently. This creates a large imbalance in the step retrieval task. For example, the \texttt{look\allowbreak\_up\_record} step is used much more often than a step in the \texttt{slack} scope, since retrieving database records is a frequent operation in workflows. We then experimented with downsampling very frequent components based on their frequency in an exponential fashion (e.g., downsample steps occurring 50 times in our dataset by 4x and downsample steps occuring 500 times by 16x). A similar procedure is applied to tables and fields although their imbalance is less drastic.

For evaluation, we use 10 splits from deployments of the enterprise system, thereby simulating evaluating on real-world customer settings. While the labeled dataset is from our internal IT domain, these other splits come from different domains and include steps and tables not present in the IT domain. Their statistics are shown in Table \ref{tab:ood_datasets}.

\begin{table}[h]
\vspace{-0.1cm}
  \centering
  \caption{Number of retrieval examples in OOD splits.}
  \vspace{-0.2cm}
  \label{tab:ood_datasets}
  \begin{tabular}{lcccc}
    \toprule
    \textbf{Dataset ID} & \textbf{Flows} & \textbf{Steps} & \textbf{Tables} & \textbf{Fields} \\
     & & \textbf{samples} & \textbf{samples} & \textbf{samples} \\
    \midrule
    OOD1  & 103 & 166 &  58 & 107 \\
    OOD2  & 100 &  94 & 209 & 392 \\
    OOD3  & 100 & 119 & 261 & 450 \\
    OOD4  & 100 & 136 & 188 & 275 \\
    OOD5  & 100 & 141 & 206 & 305 \\
    OOD6  & 100 & 179 & 255 & 529 \\
    OOD7  &  98 & 140 & 329 & 665 \\
    OOD8  & 100 & 145 & 188 & 246 \\
    OOD9  & 100 & 207 & 158 & 301 \\
    OOD10 & 171 & 134 & 477 & 690 \\
    \midrule
    \textbf{TOTAL} & \textbf{1,072} & \textbf{1,461} & \textbf{2,329} & \textbf{3,960} \\
    \bottomrule
  \end{tabular}
\end{table}

To test whether the multilingual abilities of the base model (mGTE) are preserved after our multi-task fine-tuning, we translate the development dataset using the Google Translate API.

Lastly, our aim is to generalize to similar retrieval tasks, allowing the retriever to be used in \textit{any} RAG application that needs similar kind of data. To this end, we create a task called \textit{workflow retrieval}, consisting of retrieving workflows (in YAML format) from text. This could be useful for few-shot prompting in Flow Generation. For this task, we evaluate on flows from OOD1, OOD8, and OOD9, resulting in 303 examples.

\subsection{Models}

The simplest baseline we compare against is BM25, the simplest in terms of RAG deployment. Stronger baselines are open-source multilingual models. We select the small (118M), base (278M), and large (560M) versions of mE5 \cite{wang2024multilinguale5textembeddings} to see whether more parameters help retrieval. Since mE5 has a context length of 512, we also pick mGTE-base \cite{zhang-etal-2024-mgte}, which has 8,192 context length and 305M parameters.

Our solution is to fine-tune mGTE-base using a contrastive loss objective \cite{hadsell2006dimensionality}. We prefer a retriever with large context as, once deployed, we do not control the length of instructions/context that it will receive. We also prefer a small model to allow the retrieval part of RAG to scale to many LLM concurrent calls. Appendix \ref{sec:training_details} provides details of our training process, including hyperparameters.

\subsection{Evaluation Metrics}\label{sec:metrics}

We use Recall@K \cite{manning2008introduction} as our metric across all tasks. While we briefly explored more sophisticated metrics such as Normalized Discounted Cumulative Gain (NDCG) \cite{jarvelin2002cumulated} and Mean Reciprocal Rank (MRR) \cite{voorhees1999trec}, we found that these metrics highly correlate with recall in our use case.

We report results on the most important retrieval tasks. When generating a workflow or a playbook with many steps, the LLM needs to be grounded on the customer installation of the enterprise system. In these cases, as we suggest to show many steps to the LLM, we evaluate Step@15 (K=15). When the LLM needs to include a table or field name in its generation, we suggest fewer choices, as there is typically one or two tables that are needed; hence, here we use Table@5 and Field@5 (K=5).

\section{Results}
\label{sec:results}

We first show how we arrived at our best multi-task retriever using the development set, comparing it to models fine-tuned on each task separately. We then show the OOD, multilingual, and \textit{workflow retrieval} results.

\subsection{Multi-task Retriever}
We expect multi-task fine-tuning to be better than single task fine-tuning due to a larger dataset of related tasks. However, we did not expect the imbalance in the steps distribution to cause such degradation. Table \ref{tab:single_vs_multi} shows that downsampling the data causes an improvement of 8\% in step retrieval with a small loss in field retrieval, on a fine-tuned mGTE-base model.

\begin{table}[ht]
  \centering
  \caption{Single task vs Multi-task and effect of balanced dataset. Evaluation on development split.}
  \vspace{-0.2cm}
  \label{tab:single_vs_multi}
  \begin{tabular}{lccc}
    \toprule
    \textbf{Setup} & \textbf{Step@15} & \textbf{Table@5} & \textbf{Field@5} \\
    \midrule
    Single Task         & 0.78   & 0.82 & 0.71 \\
    Multi-Task          & 0.77   & 0.86 & \textbf{0.73} \\
    + Downsampled data  & \textbf{0.86} & \textbf{0.88} & 0.71 \\
    \bottomrule
  \end{tabular}
  \vspace{-0.3cm}

\end{table}

These findings suggest that when fine-tuning a retriever for RAG, one needs to be careful on the dataset make-up. In real-world settings, there typically is a large data imbalance that needs to be handled according to the domain.

\subsection{Comparison to Baselines on OOD Splits}\label{sec:baseline_models}

Table \ref{tab:main_results} shows average results on all OOD splits, weighted by dataset size. 

\begin{table}[ht]
  \centering
  \small
  \caption{Performance of BM25, small and large open source models, and multi-task instruction fine-tuned mGTE-base (ours). Results are weighted averages across all OOD splits.}
  \vspace{-0.2cm}

  \label{tab:main_results}
  \begin{tabular}{lccc}
    \toprule
    \textbf{Model} & \textbf{Step@15} & \textbf{Table@5} & \textbf{Field@5} \\
    \midrule
    BM25        & \underline{0.82} &  \underline{0.79} &  \underline{0.26} \\
    \midrule
    mE5-Small   & 0.72 & 0.54 & 0.15 \\
    mE5-Base    & 0.74 & 0.64 & 0.15 \\
    mE5-Large   & 0.72 & 0.59 & 0.13 \\
    mE5-Large-Instruct & 0.80 & 0.74 & 0.14 \\
    mGTE-Base   & 0.72 & 0.63 & 0.08 \\
    \midrule
    GTE-Qwen2-1.5B-Instruct  & 0.21 & 0.12 & 0.01 \\
    GTE-Qwen2-7B-Instruct  & 0.18 & 0.08 & 0.03 \\
    GritLM-7B  & 0.75 & 0.59 & 0.16 \\
    E5-Mistral-7B-Instruct  & 0.57 & 0.51 & 0.16 \\

    \midrule
    Finetuned mGTE-Base (Ours)        & \textbf{0.90} & \textbf{0.90} & \textbf{0.60} \\    
    \bottomrule
  \end{tabular}
\vspace{-0.3cm}
\end{table}

On our domain-specific retrieval benchmarks, our multi-task fine-tuned mGTE-base model achieves the highest performance across all metrics, substantially outperforming both BM25 and all evaluated open-source embedding models. Notably, BM25 consistently outperforms the open-source embedding models --- regardless of model size --- including larger variants such as mE5-large and GTE-Qwen2. Increasing the scale of the mE5 retrievers from small to large yields no improvement, and in some cases results in lower scores. Only our model delivers strong, consistent gains, reaching 0.90 on both Step@15 and Table@5, and 0.60 on Field@5.

We also see that field retrieval is the most difficult task given that there is a greater variety of field names compared to tables and steps. The retriever seems to generalize well across domains as the recall numbers do not vary much from the development set (shown in Table \ref{tab:single_vs_multi}).

\subsection{Effect on Multilingual Capabilities}

We picked mGTE as our base model not only for its long context length but also because it supports multiple languages. Many open source and commercial LLMs already support multiple languages. To make RAG work properly, the retriever also needs to work well in non-English languages.

We translated the development dataset into German (DE), Spanish (ES), French (FR), Japanese (JA), and Hebrew (HE) using the Google Translate API. Table \ref{tab:results_multilingual} shows the results comparing mGTE-base and our fine-tuned version.

\begin{table}[ht]
  \centering
  \caption{Results on multilingual development datasets across step, table, and field retrieval.}
\vspace{-0.2cm}
  \label{tab:results_multilingual}
  \begin{tabular}{lcccccc}
    \toprule
    \textbf{Model} & \textbf{DE} & \textbf{ES} & \textbf{FR} & \textbf{JA} & \textbf{HE} & \textbf{Avg.} \\
    \midrule
    \multicolumn{7}{l}{\textit{Step@15}} \\
    mGTE-base & 0.53 & 0.66 & 0.66 & 0.60 & \textbf{0.47} & 0.58 \\
    Ours & \textbf{0.65} & \textbf{0.75} & \textbf{0.76} & \textbf{0.64} & 0.42 & \textbf{0.64} \\
    \addlinespace
    \multicolumn{7}{l}{\textit{Table@5}} \\
    mGTE-base & 0.34 & 0.38 & 0.42 & 0.36 & 0.39 & 0.38 \\
    Ours & \textbf{0.65} & \textbf{0.76} & \textbf{0.71} & \textbf{0.62} & \textbf{0.57} & \textbf{0.66} \\
    \addlinespace
    \multicolumn{7}{l}{\textit{Field@5}} \\
    mGTE-base & 0.15 & 0.17 & 0.16 & 0.15 & 0.17 & 0.16 \\
    Ours & \textbf{0.49} & \textbf{0.58} & \textbf{0.57} & \textbf{0.41} & \textbf{0.36} & \textbf{0.48} \\
    \bottomrule
  \end{tabular}
\vspace{-0.3cm}
\end{table}

Our multi-task fine-tuning on English datasets allows the retriever to work better than the base model on these non-English datasets. The only instance where the base model performs better is with Hebrew step retrieval. However, the average results across all three retrieval tasks are significantly lower than the English results shown in Table \ref{tab:main_results}: step retrieval goes down from 0.90 to 0.64, table retrieval from 0.90 to 0.66, and field retrieval from 0.60 to 0.48. This suggests that we may need to add some multilingual domain-specific data to the multi-task dataset.

\subsection{Generalization to \textit{Workflow Retrieval}}

Once our retriever is deployed, it can be used by any other RAG application. We therefore evaluate it on a similar task that entails retrieving workflows given text. We use recall@5 where we want to find a single workflow for a specific text. We picked three OOD splits to reduce evaluation time.

Table \ref{tab:results_workflow_retrieval} shows that the multi-task fine-tuned retriever performs better than both BM25 and the base model. But this may be too easy a task as the base model obtains 0.87 recall@5 on average. Nevertheless, our fine-tuning improved performance, showing that the multi-task dataset can transfer knowledge to similar retrieval tasks.

\begin{table}[ht]
  \centering
  \small
  \caption{Generalization on \textit{workflow retrieval} task. Metric is recall@5.}
  \label{tab:results_workflow_retrieval}
  \begin{tabular}{lcccc}
    \toprule
    \textbf{Model} & \textbf{OOD1} & \textbf{OOD8} & \textbf{OOD9} & \textbf{Avg.} \\
    \midrule
    BM25        & 0.73 & 0.60 & 0.66 & 0.66 \\    
    mGTE-base   & 0.94 & 0.76 & \textbf{0.90} & 0.87 \\
    Ours        & \textbf{0.98} & \textbf{0.93} & \textbf{0.90} & \textbf{0.94} \\
    \bottomrule
  \end{tabular}
\end{table}

\section{Conclusion}\label{sec:conclusion}

We present an approach to build a small retriever for domain-specific RAG. Via multi-task training and instruction fine-tuning, we can deploy a retriever of only 305M parameters for many applications, so that hardware costs and latency are minimized. Moreover, this retriever performs well on multilingual datasets in domain-specific use cases and shows promising results generalizing to similar retrieval tasks. Future work includes expanding the retrieval tasks and improving the multilingual capabilities of the fine-tuned model.

\clearpage

\bibliographystyle{ACM-Reference-Format}
\bibliography{main}

\appendix

\section{Training Details}\label{sec:training_details}

We fine-tuned mGTE-base for 5,000 steps, equivalent to approximately 2 epochs on the dataset. Training hyperparameters were as follows: a batch size of 32 was employed, with a per-device batch size of 2 and 16 gradient accumulation steps. Since we are not using in-batch negatives, we do not need to employ techniques such as GradCache \cite{gao2021scaling} to account for larger batch sizes. The learning rate was set to 5e-5, with a weight decay of 0.01. A warmup period of 500 steps was implemented, followed by a cosine learning rate scheduler. To optimize memory usage, gradient checkpointing was utilized \cite{chen2016training}. The Adafactor optimizer \cite{shazeer2018adafactor} was chosen for limited memory consumption.

\end{document}